\documentclass[runningheads]{llncs}

\usepackage[T1]{fontenc}

\usepackage{graphicx,verbatim}
\usepackage{booktabs}
\newcommand{\std}[1]{\text{\scriptsize$\pm$#1}}
\usepackage{amsmath}
\usepackage{amssymb}
\usepackage{xcolor}
\usepackage{pifont}
\newcommand{\xmark}{\ding{55}}
\usepackage{url}
\usepackage{graphicx}
\usepackage{subcaption}

\usepackage{color}

\usepackage{hyperref}
\begin{document}
\title{Clinical Graph-Mediated Distillation for Unpaired MRI-to-CFI Hypertension Prediction}
\titlerunning{CGMD for Unpaired MRI-to-CFI Hypertension Prediction}

\author{
Dillan Imans\inst{1} \and
Phuoc-Nguyen Bui\inst{2} \and
Duc-Tai Le\inst{3} \and
Hyunseung Choo\inst{4}\thanks{Corresponding author}
}
% index{Imans, Dillan}
% index{Bui, Phuoc-Nguyen}
% index{Le, Duc-Tai}
% index{Choo, Hyunseung}
\authorrunning{D. Imans et al.}
\institute{
Department of Computer Science and Engineering, \\ Sungkyunkwan University, Suwon, South Korea \and
Convergence Research Institute, \\ Sungkyunkwan University, Suwon, South Korea \and
Department of AI Systems Engineering, \\ Sungkyunkwan University, Suwon, South Korea \and
Department of Electrical and Computer Engineering, \\ Sungkyunkwan University, Suwon, South Korea \\
\email{\{dillanimans,phuocnguyen,ldtai,choo\}@skku.edu}
}
\maketitle

\begin{abstract}
Retinal fundus imaging enables low-cost and scalable hypertension (HTN) screening, but HTN-related retinal cues are subtle, yielding high-variance predictions. Brain MRI provides stronger vascular and small-vessel-disease markers of HTN, yet it is expensive and rarely acquired alongside fundus images, resulting in modality-siloed datasets with disjoint MRI and fundus cohorts. We study this unpaired MRI--fundus regime and introduce Clinical Graph-Mediated Distillation (CGMD), a framework that transfers MRI-derived HTN knowledge to a fundus model without paired multimodal data. CGMD leverages shared structured biomarkers as a bridge by constructing a clinical similarity $k$NN graph spanning both cohorts. We train an MRI teacher, propagate its representations over the graph, and impute brain-informed representation targets for fundus patients. A fundus student is then trained with a joint objective combining HTN supervision, target distillation, and relational distillation. Experiments on our newly collected unpaired MRI--fundus--biomarker dataset show that CGMD consistently improves fundus-based HTN prediction over standard distillation and non-graph imputation baselines, with ablations confirming the importance of clinically grounded graph connectivity. Code is available at \url{https://github.com/DillanImans/CGMD-unpaired-distillation}.

% Authors must provide keywords and are not allowed to remove this Keyword section.

\end{abstract}

\section{Introduction}
Cardiovascular diseases remain a leading cause of global morbidity and mortality, underscoring the need for scalable methods to identify individuals at elevated vascular risk early~\cite{martin20242024}. High-fidelity modalities such as brain MRI can capture downstream injury patterns associated with chronic vascular risk and small-vessel disease~\cite{hainsworth2024cerebral,sole2024impact}. However, MRI is costly and unevenly available, limiting population-scale deployment, particularly in resource-constrained settings~\cite{murali2024bringing,hricak2025strengthening}. Retinal fundus imaging, by contrast, is inexpensive and widely deployable~\cite{nakayama2025comprehensive}, but it provides an indirect and incomplete readout of vascular pathology. Hypertension (HTN)-related retinal cues are often subtle, heterogeneous, and confounded, leading to weak and high-variance signals that challenge robust inference~\cite{poplin2018prediction,zhang2020prediction,tan2022new}. This motivates a deployment-efficient strategy: train using the richer but expensive modality where available, yet require only the inexpensive, routinely acquired modality at inference.

A natural strategy is cross-modal distillation, treating MRI as teacher and fundus as student. However, most such methods assume paired multimodal data, so the teacher can provide a sample-specific target for each student image. In practice, paired MRI--fundus acquisitions are rarely available at scale: the modalities are ordered under different clinical indications, and obtaining both typically requires dedicated sub-studies rather than standard care~\cite{sun2023protocol}. This yields a disjoint multi-cohort regime in which the teacher modality is systematically missing for the student cohort, motivating methods that establish cross-cohort correspondence through shared structured clinical variables.

Related work falls into three categories. Paired cross-modal distillation~\cite{zhao2020knowledge} assumes co-acquired modalities so the teacher can supply per-instance targets; this breaks when cohorts are disjoint. Unpaired distillation~\cite{dou2020unpaired,jiang2021unpaired,yang2025umscs} transfers supervision via modality-bridging translation or reconstruction (e.g., CMEDL~\cite{jiang2021unpaired} for MRI$\rightarrow$CT segmentation), but targets closely related modalities of the same anatomy and does not address large gaps such as brain MRI$\rightarrow$fundus. Patient-similarity learning from electronic health records~\cite{gu2022structure} builds clinical similarity graphs from structured variables, but has not been used to generate patient-specific distillation targets for cross-cohort transfer.

To this end, we propose Clinical Graph-Mediated Distillation (CGMD), which transfers vascular representations from brain MRI to fundus imaging without paired acquisitions. To our knowledge, CGMD is the first distillation approach supporting cross-modal transfer across \emph{disjoint cohorts} and \emph{anatomically distinct} modalities (brain MRI$\rightarrow$fundus) without subject-level correspondence. Our main contributions are three-fold:
\begin{itemize}
    \item We identify and address an underexplored transfer setting: unpaired, disjoint-cohort MRI--fundus transfer across a large anatomical modality gap, where paired distillation is infeasible and synthesis-based alignment is ill-posed, motivating biomarker-bridged cross-cohort supervision.
    \item We propose \textbf{CGMD}, which uses shared biomarkers to build a clinical similarity graph, propagates MRI teacher embeddings, and produces patient-specific distillation targets for fundus training without paired acquisition.
    \item On a newly collected unpaired MRI--fundus--biomarker dataset, CGMD consistently improves fundus-based HTN prediction over standard distillation and non-graph imputation baselines; ablations verify that clinically grounded graph connectivity is critical for effective cross-cohort transfer.
\end{itemize}

\section{Methodology}
\begin{figure}[!t]
\centering
\includegraphics[width=0.95\textwidth]{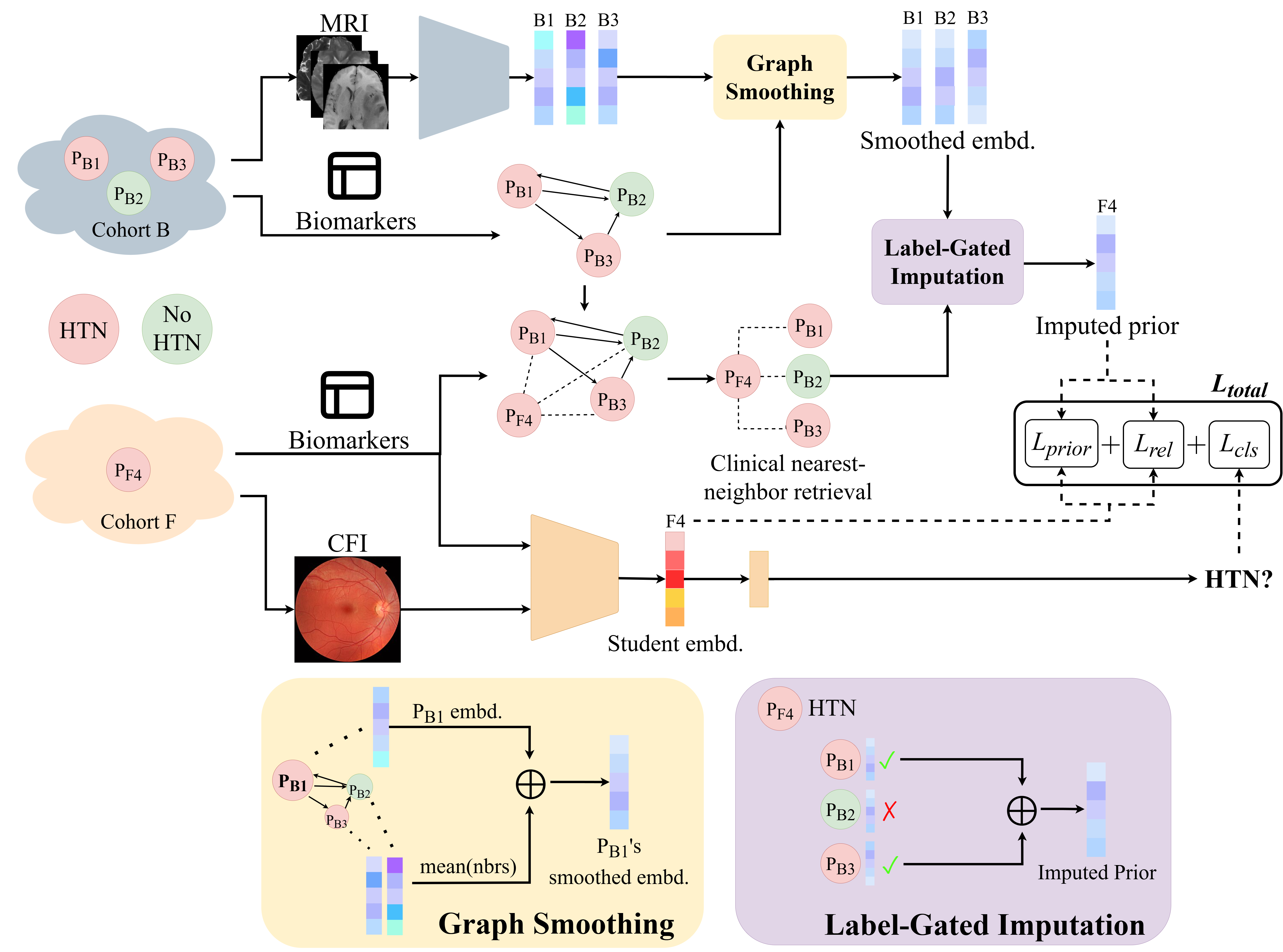}
\caption{Overview of the proposed CGMD. MRI teacher embeddings are smoothed on a clinical $k$NN graph to impute fundus priors, and the fundus student is trained with supervised, prior, and relational distillation (MRI-free inference).}
\label{fig1}
\end{figure}

Fig.~\ref{fig1} overviews our pipeline. Given disjoint MRI ($\mathcal{D}_B$) and fundus ($\mathcal{D}_F$) cohorts, we train an MRI teacher, build a $k$NN graph over MRI patients from shared clinical biomarkers, and smooth teacher embeddings into denoised priors (Sec.~\ref{sec:teacher_smoothing}). For each fundus patient we impute a patient-specific prior by label-gated aggregation over clinically nearest MRI patients (Sec.~\ref{sec:prior_imputation}), then train a fundus student with supervised, prior-distillation, and relational losses (Sec.~\ref{sec:student_training}). At inference, only fundus images and routine biomarkers are required.

\subsection{MRI Teacher and Smoothed Priors}
\label{sec:teacher_smoothing}
We first train an MRI teacher on $\mathcal{D}_B$ and extract a teacher embedding $z_i^{(0)}\in\mathbb{R}^d$ for each MRI patient $i$. If multiple MRI scans are available for a patient, we compute scan-level embeddings and average them to obtain $z_i^{(0)}$. Using the clinical biomarkers $\{c_i^B\}$, we construct a directed, weighted $k$NN graph $G_B$ by connecting each node $i$ to its $k$ nearest MRI neighbors $\mathcal{N}_B(i)$ under cosine distance in biomarker space. Here, $c_i^B\in\mathbb{R}^m$ is the patient's preprocessed clinical biomarker feature vector (numeric features concatenated with one-hot categorical variables). For each directed edge $(i\!\to\! j)$ with $j\in\mathcal{N}_B(i)$, we define the cosine distance
$d_{ij}=1-\cos(c_i^B,c_j^B)$,
convert it to a similarity weight $w_{ij}=\exp(-d_{ij}^2/\sigma^2)$, where $\sigma>0$ is a scale parameter controlling how quickly similarity decays with clinical distance, and normalize over $\mathcal{N}_B(i)$:
\begin{equation}
p_{ij}=\frac{w_{ij}}{\sum_{\ell\in\mathcal{N}_B(i)} w_{i\ell}}.
\end{equation}
We smooth teacher embeddings via one-step residual propagation:
\begin{equation}
\tilde z_i=\alpha z_i^{(0)}+(1-\alpha)\sum_{j\in\mathcal{N}_B(i)} p_{ij}\, z_j^{(0)},
\end{equation}
where $\alpha\in[0,1]$ anchors the representation to the original teacher embedding. The smoothed embeddings $\{\tilde z_i\}$ serve as teacher priors, denoising representations by borrowing signal from clinically similar patients and reducing variability from scan noise and outliers.

\subsection{Cross-cohort Prior Imputation}
\label{sec:prior_imputation}
For each fundus patient $u$, we retrieve its top-$k$ clinically nearest MRI patients $\mathcal{N}_B(u)$ by cosine distance between biomarkers $(c_u^F, c_i^B)$. Using the same distance-to-weight mapping and normalization as in Sec.~\ref{sec:teacher_smoothing}, we obtain normalized neighbor weights $\{p_{ui}\}_{i\in\mathcal{N}_B(u)}$ (defined analogously to $p_{ij}$). To enforce class-consistent transfer, we apply label gating: for a fundus patient with label $y_u^F$, we restrict retrieved neighbors to MRI patients with the same label,
\begin{equation}
\mathcal{N}_B^{+}(u)=\{\, i\in \mathcal{N}_B(u)\;:\; y_i^B = y_u^F \,\}.
\end{equation}
We define $\tilde{\mathcal{N}}_B(u)=\mathcal{N}_B^{+}(u)$ if $\mathcal{N}_B^{+}(u)\neq\emptyset$, and $\tilde{\mathcal{N}}_B(u)=\mathcal{N}_B(u)$ otherwise, and renormalize $p_{ui}$ over $\tilde{\mathcal{N}}_B(u)$. The imputed teacher prior is:
\begin{equation}
\hat z_u=\sum_{i\in \tilde{\mathcal{N}}_B(u)} p_{ui}\,\tilde z_i .
\end{equation}
Fundus data may contain multiple images per patient; we compute $\hat z_u$ once per patient and assign it to all of that patient's images.

\subsection{Student Training with Prior and Relational Distillation}
\label{sec:student_training}
We train a fundus student on $\mathcal{D}_F$ using supervised classification together with (i) distillation to imputed priors and
(ii) a relational loss defined on a clinical similarity graph over the training fundus patients.\\
\\
\textbf{Student model and classification loss.}
Given a fundus input $x_u^F$ for patient $u$, the student produces an embedding $z_u\in\mathbb{R}^d$ which is $\ell_2$-normalized and concatenated with an MLP embedding of biomarkers $h(c_u^F)$ before a single-logit prediction head. Let $\ell_u\in\mathbb{R}$ denote the predicted logit for patient $u$.
We supervise using binary cross-entropy on logits:
\begin{equation}
\mathcal{L}_{\mathrm{cls}}
=
\frac{1}{|\mathcal{B}|}\sum_{u\in\mathcal{B}}
\mathrm{BCEWithLogits}(\ell_u, y_u^F),
\end{equation}
where $\mathcal{B}$ is a minibatch of fundus patients.\\
\\
\textbf{Prior distillation.}
We match each fundus embedding to its patient-level imputed teacher prior using cosine distance. We $\ell_2$-normalize the imputed prior $\hat z_u$ before computing cosine similarity:
\begin{equation}
\mathcal{L}_{\mathrm{prior}}
=
\frac{1}{|\mathcal{B}|}\sum_{u\in\mathcal{B}}
\left(1-\cos\!\left(z_u,\hat z_u\right)\right).
\end{equation}\\
\\
\textbf{Fundus clinical graph and relational distillation.}
To constrain relations among student embeddings, we build a weighted clinical $k$NN graph $G_F=(V_F,E_F)$ over training fundus patients in biomarker space (cosine distance) and symmetrize it to obtain an undirected graph. We assign nonnegative edge weights $\pi_{uv}$ using the same distance-to-weight mapping as in Sec.~\ref{sec:teacher_smoothing}. Each patient $u$ already has a label-gated imputed prior $\hat z_u$ from Sec.~\ref{sec:prior_imputation}$;$ we $\ell_2$-normalize $\hat z_u$ when computing cosine similarities and use these priors to define teacher-implied relations for relational matching. During training, we compute the relational loss only on train-graph edges whose endpoints co-occur in the current minibatch and we gate to same-label pairs ($y_u^F=y_v^F$), yielding the within-batch edge set
\[
\mathcal{E}_{\mathcal{B}}=\{(u,v)\in E_F:\ u,v\in\mathcal{B},\ y_u^F=y_v^F\}.
\]
For each clinically similar pair $(u,v)\in\mathcal{E}_{\mathcal{B}}$ (with edge weight $\pi_{uv}$), we match the student-space similarity to the teacher-implied similarity between their priors:
\begin{equation}
\mathcal{L}_{\mathrm{rel}}
=
\frac{1}{\sum_{(u,v)\in\mathcal{E}_{\mathcal{B}}}\pi_{uv}}
\sum_{(u,v)\in\mathcal{E}_{\mathcal{B}}}
\pi_{uv}\left(
\cos(z_u,z_v)
-
\cos(\hat z_u,\hat z_v)
\right)^2,
\end{equation}
where $\cos(\cdot,\cdot)$ denotes cosine similarity. If $\mathcal{E}_{\mathcal{B}}=\emptyset$, we set $\mathcal{L}_{\mathrm{rel}}=0$. Same-label gating prevents relational constraints across disease states. Notably, $y_u^F$ enters only through loss computation---gating neighbor construction, restricting $\mathcal{L}_{\mathrm{rel}}$ to same-label pairs, and supervising $\mathcal{L}_{\mathrm{cls}}$---and never enters the student's forward pass; thus no label or graph lookup is needed at test time.\\
\\
\textbf{Overall objective.}
The training objective combines classification, prior distillation, and relational distillation:
\begin{equation}
\mathcal{L}
=
\lambda_{\mathrm{cls}}\mathcal{L}_{\mathrm{cls}}
+
\lambda_{\mathrm{prior}}\mathcal{L}_{\mathrm{prior}}
+
\lambda_{\mathrm{rel}}\mathcal{L}_{\mathrm{rel}}.
\end{equation}

\section{Experiments}

\textbf{Dataset and Experimental Setup.} We study two private cohorts from Samsung Medical Center: a brain MRI cohort (FLAIR; $n{=}295$) and a retinal fundus cohort ($n{=}112$). Cohorts are strictly disjoint at the patient level (no paired MRI--fundus subjects) and are linked only through 15 shared structured clinical variables: 7 binary indicators (sex, dyslipidemia, smoking, peripheral arterial occlusive disease, coronary artery disease (CAD), atrial fibrillation (AF), diabetes mellitus (DM)) and 8 continuous measurements (age, creatinine, blood urea nitrogen, cholesterol, triglyceride, high-density lipoprotein, low-density lipoprotein, glucose). De-identified data may be shared upon reasonable academic request, subject to institutional approval. The task is binary HTN prediction; systolic/diastolic blood pressure, explicit HTN diagnosis codes, and antihypertensive medication use are excluded from this set to prevent label leakage. While several retained variables (e.g., CAD, AF, DM) are known comorbidities of chronic HTN and may carry residual predictive signal on their own, none directly encodes blood pressure or HTN status, and we quantify this contribution explicitly via the clinical-only baseline in Fig.~\ref{fig:ablation_a}. We use 5-fold patient-level stratified cross-validation, reporting patient-level AUC/AUPRC averaged across held-out folds. Operating-point metrics (sensitivity, specificity, F1) use the fold-specific threshold maximizing Youden's $J$ on the training split and then applied unchanged to the held-out validation split. For distillation baselines that require paired samples (KD~\cite{hinton2015distilling}, FitNets~\cite{romero2014fitnets}, RKD~\cite{park2019relational}, SimKD~\cite{chen2022knowledge}), we use label-consistent surrogate pairing: each fundus sample is matched (fixed seed) to 20 randomly sampled MRI patients with the same label, and supervision uses the mean teacher output across matches; FDDM~\cite{wang2023fundus} follows its original protocol.\\
\\
\textbf{Implementation Details.} Models are implemented in PyTorch (v2.2.2, CUDA 12.1) and trained on an NVIDIA RTX 2080 Ti. The MRI teacher uses a 2D ResNet-34 with multi-slice input (16 centered slices, $256\times256$), while the fundus student uses a ResNet-18 on $224\times224$ images with a clinical vector constructed from the 15 shared variables concatenated with pooled image features before the final classifier (except in biomarker-free ablations). We train for 50 epochs with learning rate 1e-4 and batch size 8 using standard preprocessing/augmentation such as horizontal flip and normalization. Clinical graphs are cosine $k$NN ($k{=}20$ for MRI; $k{=}5$ for fundus), reflecting the relative cohort sizes ($n{=}295$ vs $n{=}112$) so the smaller fundus cohort is not given an overly diluted neighborhood. We set $\lambda_{\mathrm{cls}}=\lambda_{\mathrm{prior}}=\lambda_{\mathrm{rel}}=1$, $\sigma=1$, and $\alpha=0.9$; target quality is driven primarily by neighbor selection ($k$, label gating) rather than within-neighborhood reweighting, so near-uniform weights under $\sigma=1$ are sufficient.

% Patient-level benchmark metrics (mean ± std over K folds)
\begin{table}[b]
\centering
\caption{Patient-level comparison of distillation methods in the disjoint-cohort setting. The best results are highlighted in \textbf{bold}.}
\label{tab:kd_comparison}
\setlength{\tabcolsep}{4pt}
\begin{tabular}{lccccc}
\toprule
Method & AUC & AUPRC & Sensitivity & Specificity & F1 \\
\midrule

FDDM~\cite{wang2023fundus}
& 0.715\std{0.051}
& 0.879\std{0.013}
& \textbf{0.815}\std{0.179}
& 0.679\std{0.182}
& 0.709\std{0.032} \\

SimKD~\cite{chen2022knowledge}
& 0.732\std{0.134}
& 0.869\std{0.075}
& 0.705\std{0.234}
& 0.786\std{0.192}
& 0.761\std{0.155} \\

RKD~\cite{park2019relational}
& 0.756\std{0.074}
& 0.880\std{0.062}
& 0.651\std{0.176}
& 0.871\std{0.194}
& 0.752\std{0.104} \\

FitNets~\cite{romero2014fitnets}
& 0.779\std{0.058}
& 0.892\std{0.049}
& 0.716\std{0.108}
& 0.881\std{0.109}
& 0.807\std{0.060} \\

KD~\cite{hinton2015distilling}
& 0.803\std{0.101}
& 0.905\std{0.065}
& 0.770\std{0.151}
& 0.843\std{0.183}
& \textbf{0.832}\std{0.084} \\

\textbf{CGMD}
& \textbf{0.855}\std{0.127}
& \textbf{0.937}\std{0.055}
& 0.728\std{0.123}
& \textbf{0.933}\std{0.149}
& 0.826\std{0.087} \\
\bottomrule
\end{tabular}
\end{table}

\section{Performance Evaluation}
\textbf{Comparison with KD Methods.} Table~\ref{tab:kd_comparison} compares CGMD with distillation baselines adapted to our disjoint-cohort setting across AUC, AUPRC, sensitivity, specificity, and F1. Among the baselines, logit-based KD is strongest (AUC 0.803, AUPRC 0.905, F1 0.832), while FitNets and RKD are competitive but lower on AUC/AUPRC. CGMD improves AUC by 0.052 and AUPRC by 0.032 over the best baseline, with the highest specificity (0.933) and comparable F1 (0.826 vs.\ 0.832), but at the cost of lower sensitivity (0.728 vs.\ 0.770 for KD and 0.815 for FDDM), reflecting an operating point that favors specificity over sensitivity. These gains indicate that clinically grounded graph-based imputation provides more informative teacher targets than label-matched surrogate pairing, though the sensitivity trade-off should be weighed alongside the ranking-based gains.

\begin{table}[h]
\centering
\caption{Module ablation for CGMD using patient-level metrics. The best results are highlighted in \textbf{bold}.}
\label{tab:module_ablation}
\setlength{\tabcolsep}{2pt}
\begin{tabular}{ccccccccc}
\toprule
Distill & Smooth & Rel & AUC & AUPRC & Sensitivity & Specificity & F1 \\
\midrule

\checkmark & \xmark & \xmark
& 0.830\std{0.133}
& 0.927\std{0.060}
& 0.766\std{0.202}
& 0.876\std{0.137}
& 0.831\std{0.136} \\

\checkmark & \checkmark & \xmark
& 0.844\std{0.120}
& 0.925\std{0.066}
& 0.714\std{0.158}
& \textbf{0.933}\std{0.149}
& 0.813\std{0.115} \\

\checkmark & \xmark & \checkmark
& 0.815\std{0.116}
& 0.907\std{0.064}
& \textbf{0.820}\std{0.151}
& 0.790\std{0.170}
& \textbf{0.853}\std{0.090} \\

\checkmark & \checkmark & \checkmark
& \textbf{0.855}\std{0.127}
& \textbf{0.937}\std{0.055}
& 0.728\std{0.123}
& \textbf{0.933}\std{0.149}
& 0.826\std{0.087} \\

\bottomrule
\end{tabular}
\end{table}

\noindent\textbf{Ablation Studies.} Table~\ref{tab:module_ablation} breaks down the contribution of each CGMD component. Prior distillation alone provides a solid baseline, while adding MRI-graph smoothing improves AUC with comparable AUPRC, supporting the hypothesis that smoothing denoises teacher embeddings before cross-cohort imputation. Using the relational loss without smoothing shifts the operating point toward higher sensitivity and F1 but with a drop in AUC. Combining distillation, smoothing, and relational matching achieves the best overall AUC/AUPRC and restores balanced performance, indicating that smoothing stabilizes the imputed targets so relational constraints become beneficial rather than over-regularizing. Overall, the ablations suggest that clinical-graph smoothing is the key enabler that makes relational distillation effective in the disjoint-cohort regime.

% we can put like bars for the data and prior building here just rq

\begin{figure}[h]
    \centering
    \begin{subfigure}[t]{0.475\linewidth}
        \centering
        \includegraphics[width=\linewidth]{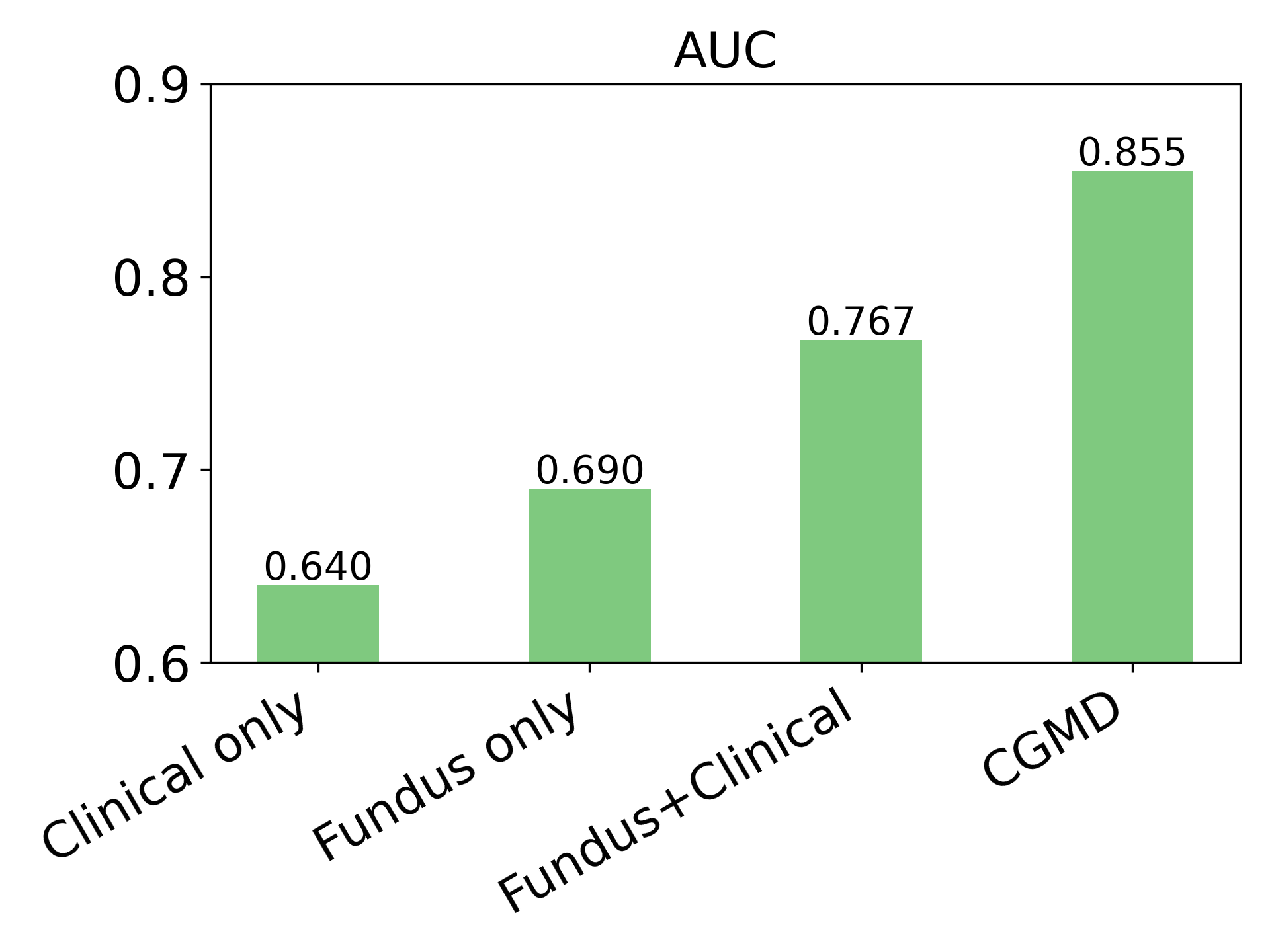}
        \caption{Data availability ablation.}
        \label{fig:ablation_a}
    \end{subfigure}
    \hfill
    \begin{subfigure}[t]{0.475\linewidth}
        \centering
        \includegraphics[width=\linewidth]{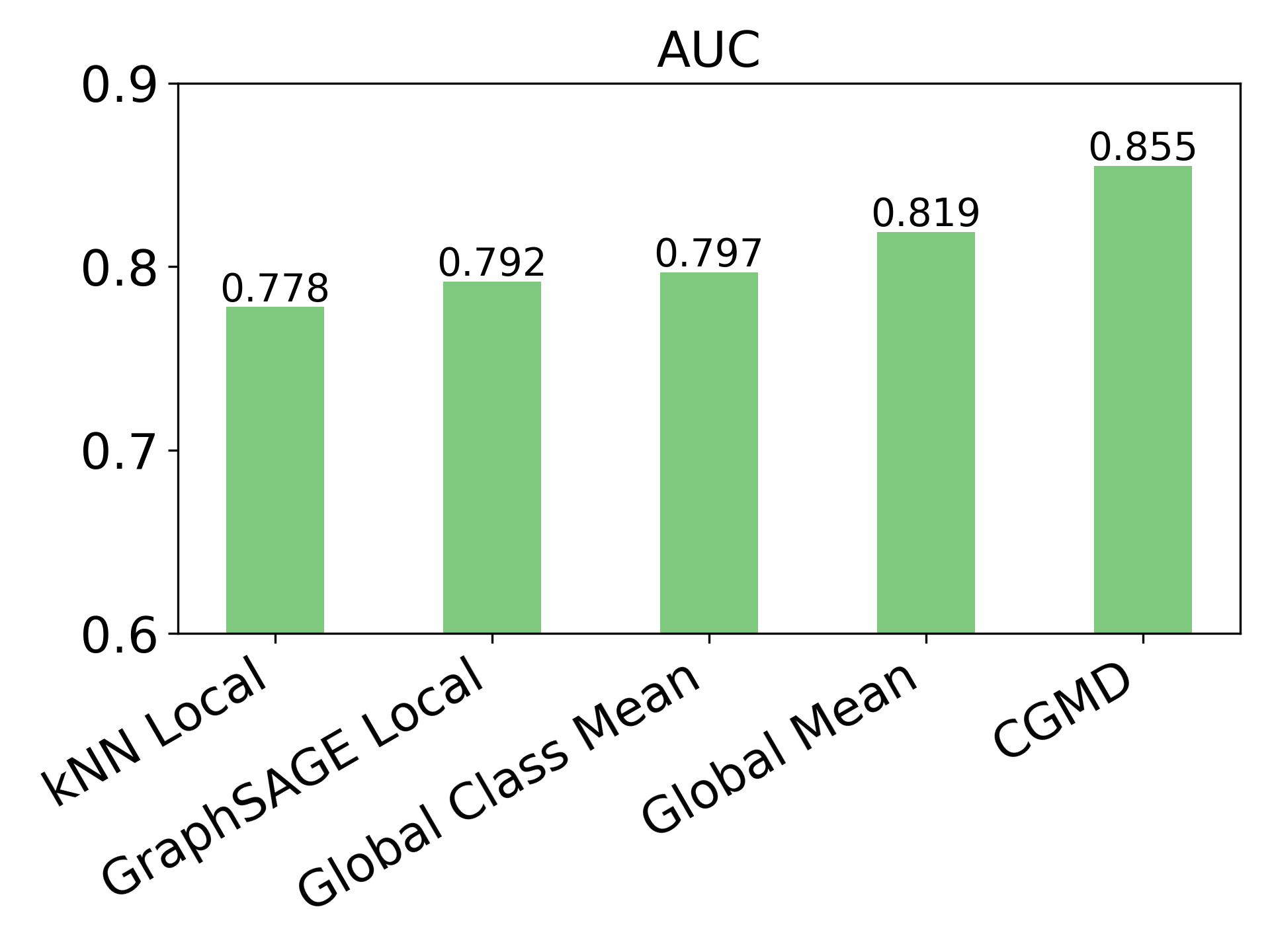}
        \caption{Prior construction ablation.}
        \label{fig:ablation_b}
    \end{subfigure}

    \caption{Impact of (a) data availability and (b) prior construction strategies on the final HTN classification performance.}
    \label{fig:ablation}
\end{figure}

Fig.~\ref{fig:ablation_a} ablates data availability: performance improves as more informative inputs are provided, and CGMD remains best overall, indicating that MRI$\rightarrow$fundus adds benefit beyond clinical features and/or fundus images alone (or simple concatenation). Critically, the Fundus+Clinical condition already accesses the same 15 biomarkers used to build CGMD's targets, so the further gain to CGMD is attributable solely to the MRI-derived graph-mediated supervision, holding biomarker access fixed and ruling out a clinical-feature shortcut. Fig.~\ref{fig:ablation_b} ablates prior construction for the unpaired fundus cohort. Local baselines propagate teacher embeddings over a clinical similarity graph ($k$NN; GraphSAGE~\cite{hamilton2017inductive}) without label gating, while global baselines use class-conditional means (Global Class Mean) or cohort-level averages (Global Mean). CGMD outperforms all alternatives, consistent with the benefit of combining patient-specific, graph-mediated priors with class-consistent (label-gated) aggregation; however, as the local baselines also differ in propagation strategy, this comparison does not isolate label gating alone, and a controlled ablation is left to future work.

\begin{figure}[!h]
    \centering
    \includegraphics[width=0.9\textwidth]{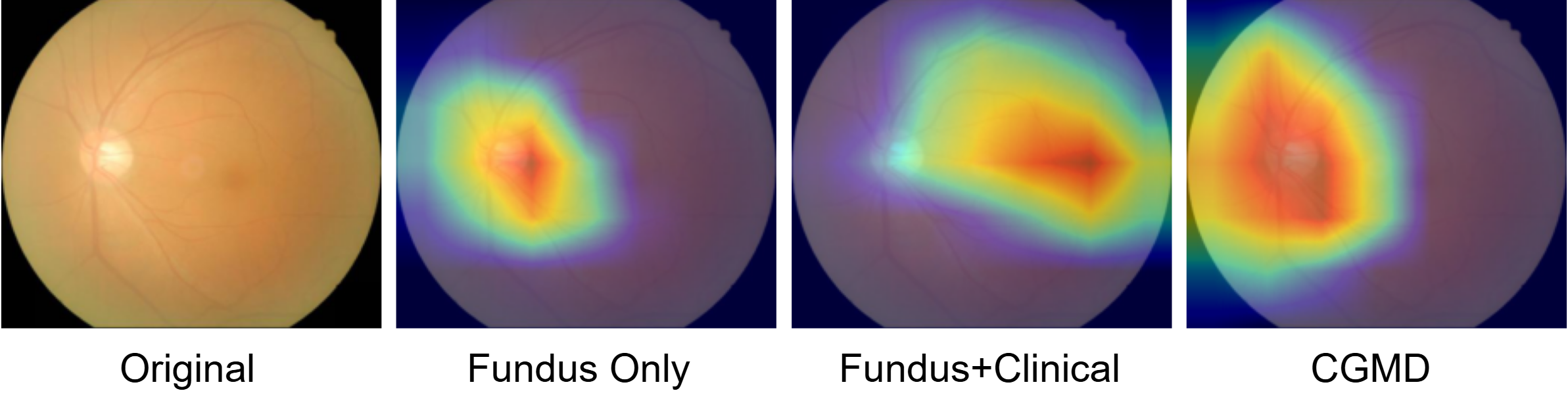}
    \caption{Grad-CAM visualizations for a representative disagreement case.}
    \label{fig:gradcam}
\end{figure}
\noindent\textbf{Qualitative Analysis.} Fig.~\ref{fig:gradcam} shows a representative disagreement case in which CGMD predicts correctly while both baselines fail. The baselines concentrate on a narrow region near the optic disc, whereas CGMD attends to a broader retinal area, consistent with diffuse vascular feature utilization.

\section{Conclusion}
We presented Clinical Graph-Mediated Distillation (CGMD), a resource-conscious framework for disjoint-cohort cross-modal transfer that uses shared biomarkers to propagate MRI teacher embeddings as patient-specific distillation targets for fundus training requiring neither paired acquisitions nor MRI at deployment. On 5-fold cross-validation, CGMD outperformed adapted distillation baselines, with ablations confirming the importance of clinically grounded graph structure. As a single-center study with a modest fundus cohort ($n{=}112$), these results are preliminary; future work will pursue external validation on larger multi-center cohorts, extending to other disjoint-cohort modality pairs.

\begin{credits}
\subsubsection{\ackname} This work was supported in part by the Korea government (MSIT), IITP, under IITP-2026-RS-2020-II201821 (60\%) and RS-2019-II190421 (10\%); and by the Ministry of SMEs and Startups (MSS) under RS-2024-00514724 (30\%). Professors D.T. Le and H. Choo are also with SKAI X Inc.

\subsubsection{\discintname}The authors have no competing interests to declare that are relevant to the content of this article.
\end{credits}

\bibliographystyle{splncs04}
\bibliography{references}




% LNCS Homepage, \url{http://www.springer.com/lncs}, last accessed 2023/10/25
% \end{thebibliography}

\end{document}